%% file: main.tex
\documentclass[letterpaper, 10 pt, conference]{ieeeconf}  

\IEEEoverridecommandlockouts                              

\usepackage{graphics} 
\usepackage{epsfig} 
\usepackage{mathptmx} 
\usepackage{times} 
\usepackage{amsmath} 
\usepackage{amssymb}  
\usepackage{multirow}
\usepackage{color,soul}
\usepackage{algorithm}
\usepackage{siunitx}
\usepackage{algpseudocode}
\usepackage{caption}
\usepackage{subcaption}
\usepackage{xcolor}

\usepackage{hyperref}
\usepackage[comma,numbers]{natbib}
\title{\LARGE \bf
Efficient, Dynamic Locomotion through Step Placement with Straight Legs and Rolling Contacts}

\author{Stefan Fasano$^{1}$,
James Foster$^{1,2}$,
Sylvain Bertrand$^{1}$,
Christian DeBuys$^3$,
and Robert Griffin$^{1,2,3}$
\thanks{This work was funded through ONR Grant N00014-19-1-2023, NASA Grant  80NSSC20M0197, and DAC Cooperative Agreement W911NF-21-2-0241.}
\thanks{$^{1}$Author with the Institute for Human and Machine Cognition}
\thanks{$^{2}$Author with the University of West Florida}
\thanks{$^{3}$Author with Boardwalk Robotics, Inc.}
\thanks{Email : \url{{sfasano, jfoster, sbertrand, rgriffin}@ihmc.org}, \url{christian.debuys@boardwalkrobotics.com}
}} 

\begin{document}

\maketitle
\thispagestyle{empty}
\pagestyle{empty}

\begin{abstract}
For humans, fast, efficient walking over flat ground represents the vast majority of locomotion that an individual experiences on a daily basis, and for an effective, real-world humanoid robot the same will likely be the case. In this work, we propose a locomotion controller for efficient walking over near-flat ground using a relatively simple, model-based  controller that utilizes a novel combination of several interesting design features including an ALIP-based step adjustment strategy, stance leg length control as an alternative to center of mass height control, and rolling contact for heel-to-toe motion of the stance foot. We then present the results of this controller on our robot Nadia, both in simulation and on hardware. These results include validation of this controller's ability to perform fast, reliable forward walking at 0.75 m/s along with backwards walking, side-stepping, turning in place, and push recovery. We also present an efficiency comparison between the proposed control strategy and our baseline walking controller over three steady-state walking speeds. Lastly, we demonstrate some of the benefits of utilizing rolling contact in the stance foot, specifically the reduction of necessary positive and negative work throughout the stride.
\end{abstract}

\section{Introduction}
\label{sec:introduction}
\input{introduction}

\section{Background}
\label{sec:background}
\input{background}

\section{Controller Design}
\label{sec:controller_design}
\input{controller_design}

\section{Results}
\label{sec:results}
\input{results}

\section{Conclusion and Future Work}
\label{sec:conclusion}
\input{conclusion}

\section{Acknowledgements}

We would like to thank Nick Kitchel for assistance in experimentation and William Howell for capturing the videos for this work. Also thanks to the mechanical team for keeping the robot in good running condition.

\newpage
\pagebreak

\bibliography{mybib}

\end{document}

%% file: introduction.tex
The ability to walk quickly and robustly is a fundamental requirement for legged robots to be useful in real-world tasks.
While much of the promise of humanoid robots lies in their potential ability to traverse complex, discontinuous surfaces, they must also be capable of traveling over relatively flat ground in a quick and efficient manner. Humans perform both locomotion styles very well, with the ability to pick out and execute specific contact sequences when walking over complex terrain, while equally capable of walking down a hallway or over fields with little thought given to the precise contact location.
While much of our previous work has centered around the former \cite{Koolen_2016, griffin2019footstep}, in this paper we present a controller designed to quickly, efficiently, and blindly walk over relatively flat ground.

Stable bipedal walking is a challenging problem due to the high dimensionality, underlying nonlinearity, limited control authority, and hybrid nature of the locomotion dynamics. 
The dominant approach has thus been the use of reduced-order and simplified models, most notably with the linear inverted pendulum (LIP) model \cite{garcia1998simplest, kajita20013d} and its permutations like zero-moment point \cite{kajita2003biped}, capture point \cite{pratt2006capture, pratt2012capturability}, and divergent component of motion \cite{englsberger2013three}. More generally, gross robot quantities such as the Center of Mass (CoM) position and velocity \cite{boroujeni2021unified} or the net angular momentum \cite{herzog2016momentum} are controlled with inverse dynamics solvers used to generate joint-level commands from these quantities \cite{Koolen_2016, kuindersma2016optimization, feng2015optimization}. A commonly cited issue with these models, though, is that some of the simplifying assumptions tend to reduce the efficiency of the gait, and omit salient features of the dynamics.

\begin{figure}[t!]
\centering
    \includegraphics[width=0.875\columnwidth]{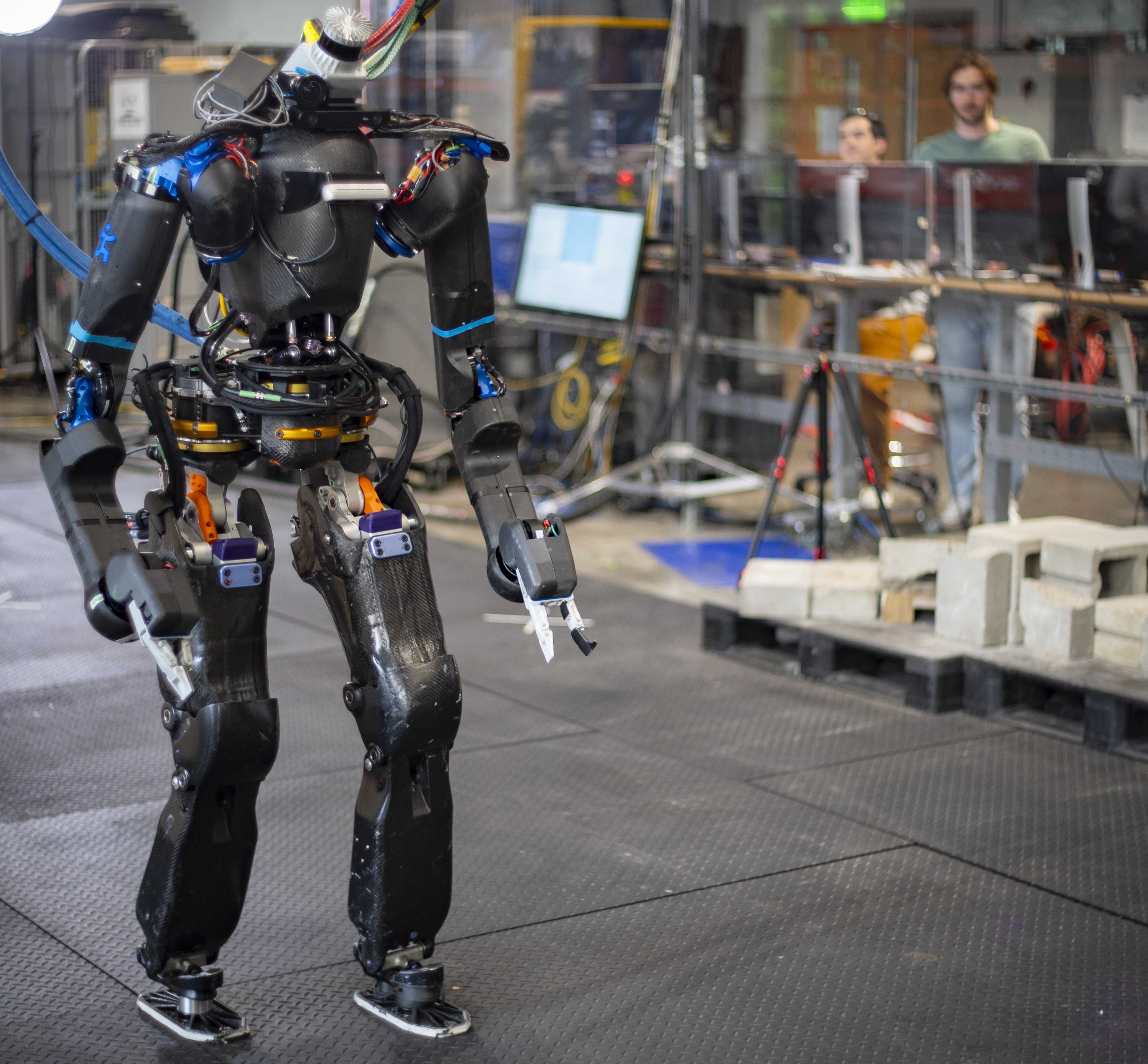}
    \vspace{-0mm}
    \caption{Our humanoid robot, Nadia, during forward walking hardware tests of the Quickster walking controller.}
    \vspace{-6mm}
    \label{fig:nadia_fast_walking}
\end{figure}

To address these limitations, alternatives have been utilized, such as Hybrid Zero Dynamics (HZD) \cite{westervelt2003hybrid, grizzle2014models}, where nonlinear optimization is used to design gaits as periodic orbits in the state space of a hybrid dynamics model. 
However, in online applications, HZD approaches typically require the introduction of additional control heuristics, and have a limited set of available motions due to the reliance on the offline library of periodic gaits \cite{reher2016realizing}.
In a similar vein, trajectory optimization has shown very promising results, especially for long-horizon, complex motion generation \cite{mastalli2020crocoddyl, ponton2021efficient, posa2014direct}. Similar to HZD, though, these motions are typically limited by their reliance on offline computation. The real-time equivalent, whole-body model predictive control (MPC) \cite{dantec2022whole, galliker2022planar}, is emerging as an attractive option for dynamic locomotion, but is often stymied in achieving real-time control rates by the computational complexity of the underlying optimal control problem.

A recently proposed approach that leverages the simplicity and other benefits of traditional LIP model-based controllers while including a more complete picture of the system dynamics is the Angular Linear Inverted Pendulum (ALIP) model {\cite{gong2021one}}. The ALIP model encodes both the CoM velocity and angular momentum into the system model, making controllers that employ this model less susceptible to deviations from the true system dynamics, particularly in the case of robots with higher distal mass like our 29 degree of freedom humanoid Nadia (see Fig. {\ref{fig:nadia_fast_walking}}) performing quicker walking with faster swings. The use of this model in step adjustment-based walking controllers such as in {\cite{gong2021one}} has been promising, but there remains an opportunity to reconcile the benefits of this model with solutions that address the inefficiencies that result from its simplifying assumptions.


In this paper, we present Quickster; a simple, model-based locomotion controller designed for fast, efficient walking. Quickster seeks to expand on the work in {\cite{gong2021one}}  by augmenting the ALIP-based step adjutment strategy with two additional design features aimed at further improving the performance and efficiency of locomotion: (1) To control the height of the robot, we choose to regulate only the stance leg length, rather than the CoM height directly, allowing walking with straighter legs than the classical LIP gaits \cite{griffin2018straight}. (2) To further increase efficiency, we introduce a heel-to-toe rolling contact motion of the Center of Pressure (CoP), which has been observed in biological systems to increase walking efficiency \cite{adamczyk2006advantages}, and can be passively achieved using musculoskeletal models {\cite{narioka20093d}}, but has yet to be demonstrated in an actively controlled context. Using this controller, Nadia can walk stably at 0.75 m/s and recover from external disturbances. The main contributions of this paper are:
\begin{itemize}
    \item Formulation of an ALIP-based step adjustment controller with strategically shaped feedback dynamics.
    \item Strategy for directly controlling the stance leg length, and exploration of how this affects the validity of the ALIP model behind the Quickster controller.
    \item Introduction of a novel rolling contact strategy based on biological system observations, and investigation of the efficiency improvements this approach provides.
    \item Validation of this unique combination of design features on a full-size humanoid robot, along with a performance and efficiency comparison against a baseline, center-of-mass trajectory-based capture point controller {\cite{griffin2023reachability}}.
\end{itemize}


%% file: background.tex
The LIP model represents the robot as a point mass that moves at a constant height above the ground, and exerts all its forces through a point foot. This leads to the CoM dynamics:
\begin{equation}
    \ddot{\mathbf{x}} = \frac{g}{\Delta z}\left( \mathbf{x} - \mathbf{r}_{cop} \right),
    \label{eqn:lip_dynamics}
\end{equation}
where $\mathbf{x}$ is the CoM position in $x$-$y$, $\mathbf{r}_{cop}$ is the CoP position in $x$-$y$, $g$ is gravity, and $\Delta z$ is the height of the CoM above the ground. This is often written assuming no height acceleration as $\ddot{\mathbf{x}} = \omega^2 \left( \mathbf{x} - \mathbf{r}_{cop} \right)$, where $\omega = \sqrt{\frac{g}{\Delta z}}$ is the natural frequency of the inverted pendulum. From these dynamics, we can find the solution as
\begin{equation} 
\left[ \begin{array}{c} 
\mathbf{x}(T) \\ \dot{\mathbf{x}}(T)
\end{array}\right] 
= 
\left[ \begin{array}{cc}
c \left(\omega \Delta T \right) & \tfrac{1}{\omega} s \left(\omega \Delta T  \right)
\\
\omega s \left( \omega \Delta T \right) &  c \left(\omega \Delta T \right)
\end{array} \right]
\left[ \begin{array}{c} 
\mathbf{x}(t) \\ \dot{\mathbf{x}}(t)
\end{array}\right],
\label{eqn:lip_step_to_step}
\end{equation}
where $\Delta T = T - t$, $c$ and $s$ are $cosh$ and $sinh$, respectively, and we assume $\mathbf{r}_{cop}$ is the origin.

The ALIP model proposes an alternative rate state, where the angular momentum about the contact point, $L_y$ for the $x$ component, is used \cite{gong2021one}.
This then changes the dynamics to
\begin{equation}
    \left[ \begin{array}{c} 
    \dot{x} \\ \dot{L}_y
    \end{array} \right]
    =
    \left[ \begin{array}{cc} 
    0 & \frac{1}{m\Delta z} \\
    mg & 0
    \end{array} \right]
        \left[ \begin{array}{c} 
    x \\ L_y
    \end{array} \right],
    \label{eqn:alip_dynamics}
\end{equation}
which has the solution
\begin{equation} 
\left[ \begin{array}{c} 
x(T) \\ L_y(T)
\end{array}\right] 
= 
\left[ \begin{array}{cc}
c\left( \omega \Delta T \right) & \tfrac{1}{m \Delta z\omega} s\left(\omega \Delta T \right)
\\
m \Delta z \omega s\left(\omega \Delta T \right) &  c\left(\omega \Delta T \right)
\end{array} \right]
\left[ \begin{array}{c} 
x(t) \\ L_y(t)
\end{array}\right].
\label{eqn:alip_step_to_step}
\end{equation}
This formulation has the distinct advantage of considering the angular momentum about stance foot, which allows incorporation of any angular momentum about the CoM into the state estimate, breaking the assumption of a point mass present in the LIP model. 
The generation of angular momentum is required for the robot to swing its leg, but can typically be ignored due to low swing speeds or low leg inertia. 
For our robot to achieve faster walking speeds, however, this is not the case, so consideration of angular momentum in the robot dynamics is essential. 
As shown in \cite{gong2021one}, this ALIP model provides a better estimate of the velocity at the end of the step when angular momentum is generated, which will greatly improve the performance of controllers based on step placement.

%% file: controller_design.tex
The main purpose of the Quickster controller is to regulate foot placement to achieve stable walking at a desired speed, while also increasing efficiency of locomotion. This can be done by first using step adjustment to achieve stable, closed-loop step-to-step dynamics, with efficiency added through an alternative method for controlling the robot's height, while also using heel-to-toe motions of the CoP. These feedback and feedforward objectives are then fed into a whole-body inverse dynamics-based  quadratic program (QP) to determine joint-level commands \cite{Koolen_2016}. 

\subsection{Step Placement Control}

\begin{figure*}[t!]
\centering
    \includegraphics[width=0.86\textwidth]{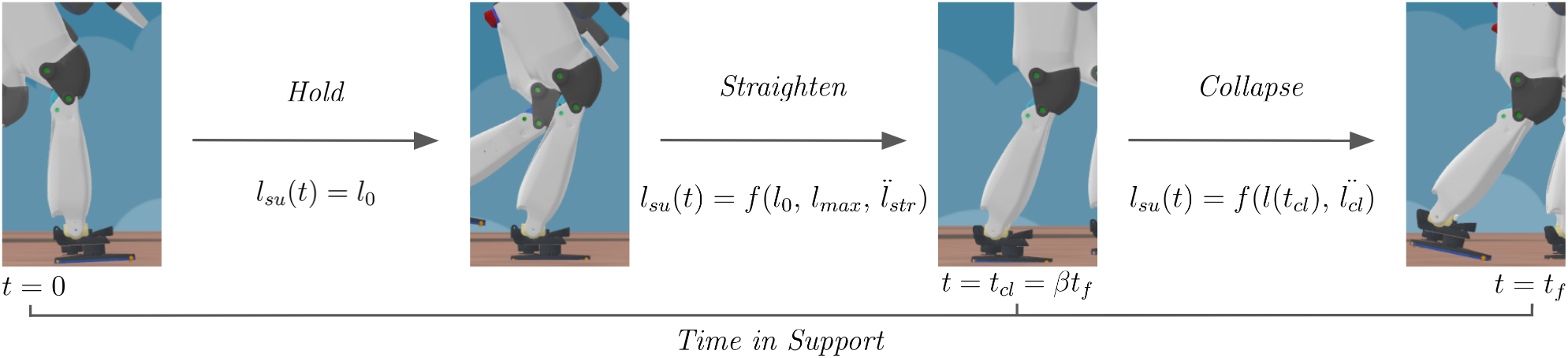}
    \vspace{-1.25mm}
    \caption{Stages of stance leg length control. Leg length is initially held constant, then increases with a constant acceleration towards a maximum length, and finally decreases with a constant acceleration to motivate touchdown of the swing foot.}
    \vspace{-6mm}
    \label{fig:leg_length}
\end{figure*}

Capture point dynamics {\cite{pratt2006capture}} dictate that the capture point strays further from the stance foot at higher walking speeds, decreasing the control authority that ankle-based walking strategies have over the CoM dynamics. For this reason, Quickster employs a step adjustment-based strategy. This approach also provides a built-in capability for push recovery, as the desired touchdown position adjusts throughout swing based on the CoM dynamics. While previous works using the ALIP model for step placement have designed a deadbeat controller such that the robot returns to the desired velocity in a single step \cite{gong2021one}, or use a simple MPC to do so in multiple steps \cite{gibson2022terrain}, in this work we prefer to specify control gains by shaping the feedback dynamics.
We can observe that the robot is able to change the dynamics in Eq. \ref{eqn:alip_step_to_step} through selection of the step position relative to the CoM with the following feedback law (forward direction shown):
\begin{equation}
    x_{k+1} = \frac{-k_p}{m \Delta z}L_{y,k+1}
\label{eqn:step-placement-law},
\end{equation}
which is akin to predicting the next touchdown position based on the centroidal velocity using a Raibert heuristic {\cite{raibertHeuristic1986}\cite{raibertHeuristic1991}.
Augmenting Eq. {\ref{eqn:alip_step_to_step}} with Eq. {\ref{eqn:step-placement-law}} makes the closed loop ALIP dynamics}
\begin{equation} 
\left[ \begin{array}{c} 
x_{k+1} \\ L_{y,k+1}
\end{array}\right] 
= 
\left[ \begin{array}{cc}
-k_p \omega s\left(\omega T \right) &  -\frac{k_p}{m \Delta z} c\left(\omega T \right)
\\
m \Delta z \omega s\left(\omega T \right) &  c\left(\omega T \right)\end{array} \right]
\left[ \begin{array}{c} 
x_k \\ L_{y,k}
\end{array}\right].
\label{eqn:closed-loop-alip}
\end{equation}
From here, we employ pole placement to select $k_p$ such that the desired closed-loop eigenvalue, $\lambda$, is achieved: 
\begin{equation} 
 k_p = \frac{c(\omega T) -  \lambda}{m \Delta z \omega s(\omega T)}.
\label{eqn:pole-placement}
\end{equation}
Our selection of $\lambda$ allows us to distribute balance recovery from step-to-step, achieving the same deadbeat performance as \cite{gong2021one} if $\lambda = 0$. 

While the controller in Eq. {\ref{eqn:step-placement-law}} ensures convergence to a stable zero velocity state at a rate defined by $\lambda$, to enable the robot to walk with a given velocity and step width, we need to add offsets to our desired footstep. 
This can be derived from the capture point dynamics \cite{pratt2006capture} as
\begin{equation} 
\Delta \xi_{sw} = \frac{\mu}{1 + e^{\omega T}}
\quad\mathrm{and}\quad 
\Delta \xi_{sp}(v_d) = - \frac{v_d T}{e^{\omega T} - 1},
\label{eqn:step-offset}
\end{equation}
where $\mu$ is the desired step width and $v_d$ is the desired speed in $x$ or $y$. Combining Eqs. {\ref{eqn:step-placement-law}}, \ref{eqn:closed-loop-alip}, \ref{eqn:pole-placement}, and \ref{eqn:step-offset} for both the $x$ and $y$ directions, we can derive our step placement control law that will enable the robot to walk with the forward velocity, lateral velocity, and step width that we desire:
\begin{equation} 
    \mathbf{x}_{k+1} = \begin{bmatrix}
    -k_p \omega x_k s\left(\omega T \right) - \frac{k_p}{m\Delta z} L_{y,k} c\left(\omega T \right)
- \Delta \xi_{sp}(\dot{x}_d)
    \\
    -k_p  \omega y_k s\left(\omega T \right)  + \frac{k_p}{m\Delta z} L_{x,k} c\left(\omega T \right)
    - \Delta \xi_{sp}(\dot{y}_d) - \Delta \xi_{sw}
    \end{bmatrix}.
\label{eqn:desired_step}
\end{equation}

\subsection{Height Control through Leg Length}

\begin{figure} [b!]
\vspace{-5mm}
    \centering
     \begin{subfigure}[b]{0.49\columnwidth}
         \centering
         \includegraphics[width=\columnwidth]{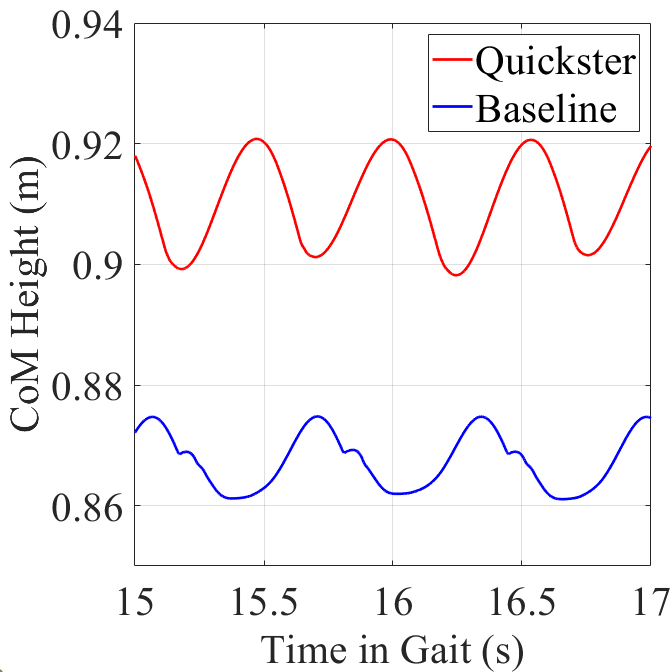}
         \caption{CoM height vs. time}
         \label{fig:com_height}
     \end{subfigure}
     \hfill
     \begin{subfigure}[b]{0.49\columnwidth}
         \centering
         \includegraphics[width=\columnwidth]{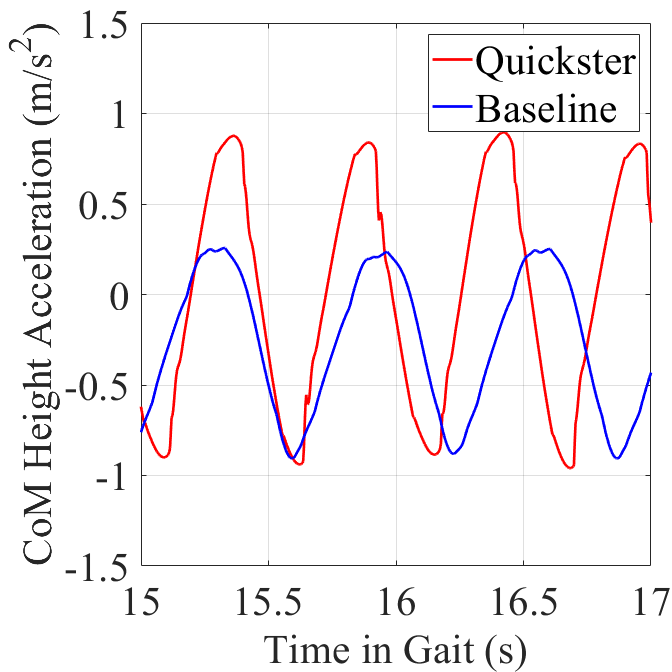}
         \caption{CoM height acceleration vs. time}
         \label{fig:com_height_acceleration}
     \end{subfigure}
    \vspace{-4mm}
    \caption{CoM height and height acceleration comparison between controllers at 0.7 m/s steady-state walking.}
    \label{fig:height_and_height_acceleration_comparison}
\end{figure}

A common technique of many bipedal walking controllers is the direct control of the CoM height dynamics \cite{englsberger2013three, ficht2022direct}.
Often, this is done to satisfy many of the assumptions of the underlying simplified models described in Sec. \ref{sec:background}, however, kinematic constraints require the variation of height when taking longer steps with straighter legs, both of which are required for more efficient walking \cite{griffin2017capture}.
Additionally, by controlling CoM height rather than leg length, the controller is susceptible to singularities when the knee is too straight.
This can be addressed through complex CoM height planning,  which often leads to nonlinear formulations \cite{huang2021knee, park2023heel, dafarra2020non}.
Instead, to simplify height control over varied terrain and gain the efficiencies from walking with straighter legs, we propose an alternative paradigm where desired length changes of the support leg are controlled, rather than the CoM height directly, similar to our previous work {\cite{griffin2018straight}}.

Our implementation of indirect walking height modulation through support leg length control can be seen in Fig. \ref{fig:leg_length}. 
On touchdown, the stance leg enters a hold state in which its length remains constant at its initial value $l_0$ in order to prevent undesirable horizontal CoM deceleration due to premature leg straightening. 
Once the CoM is within a defined distance of the stance foot, the support leg will begin to straighten to a desired maximum length $l_{max}$ at a constant  acceleration $\ddot{l}_{str}$. The leg will either continue straightening, or remained fully straightened, until a fraction of total stance time $\beta$ is reached, at which point the leg will collapse at a constant acceleration $\ddot{l}_{cl}$. The collapse state serves to counteract the vertical component of the toe-off motion, while encouraging touchdown of the swing foot in the event of longer steps.


To justify our choice of allowing CoM height fluctuations despite the use of a simplified ALIP model, we analyzed the CoM height and CoM height acceleration of the Quickster controller, and compared them against those of the baseline controller in Fig. \ref{fig:height_and_height_acceleration_comparison}. 
The CoM height of both controllers have remarkably similar profiles, shown in Fig. \ref{fig:com_height}, despite their different approach to  height control. 
This plot also highlights the higher walking height achieved by Quickster's stance leg length control strategy. 
More important to the assumptions associated with the ALIP model, however, is the CoM height acceleration, shown in Fig. \ref{fig:com_height_acceleration}. Both controllers experience similar, sufficiently low height acceleration (less than 0.1g) during steady-state walking such that the assumption of constant CoM height can still reasonably hold.

\subsection{Definition of Rolling Contact}

\begin{figure}[b!]
\centering
    \vspace{-3mm}
    \includegraphics[width=0.5\columnwidth]{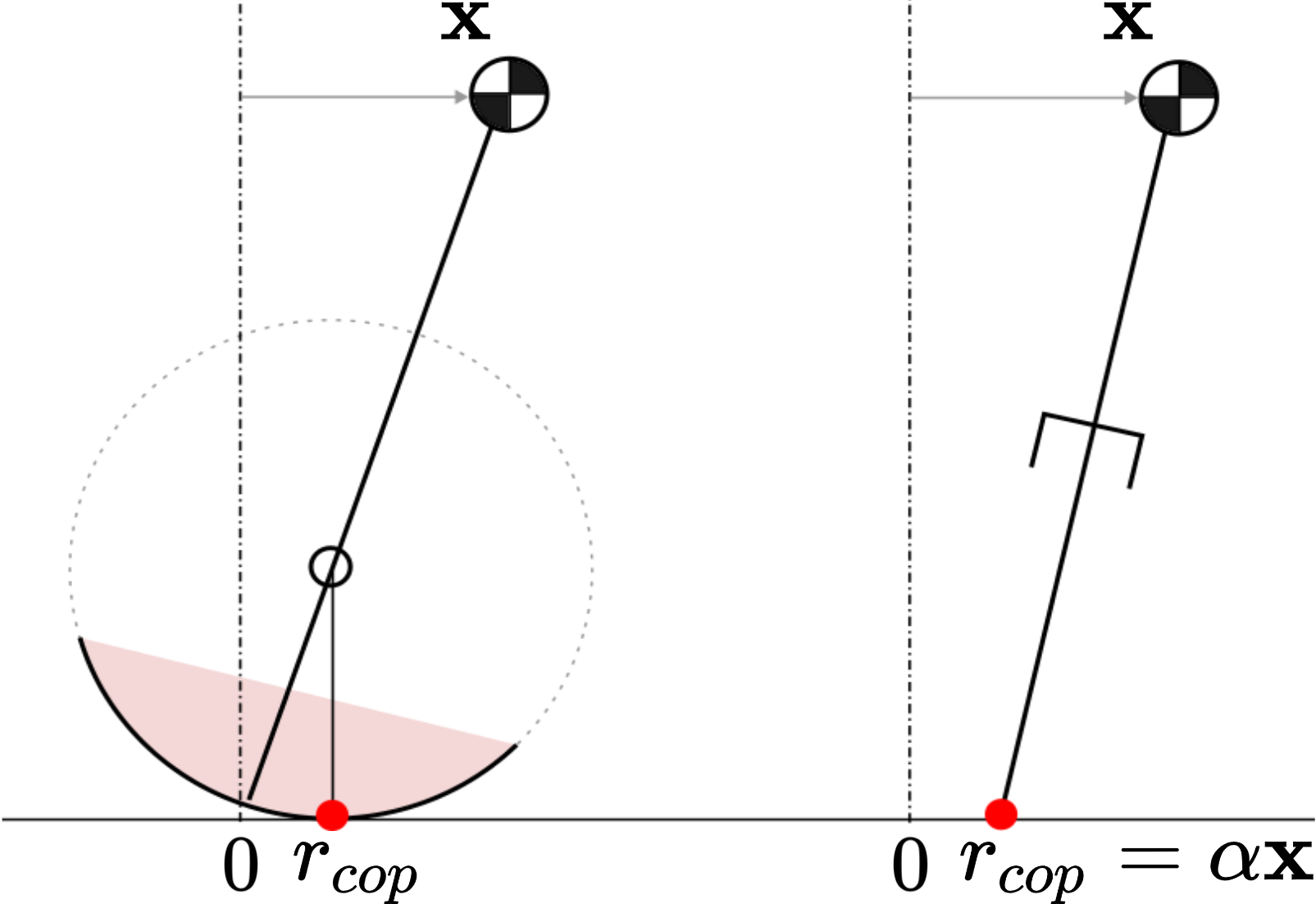}
    \caption{Left: biological walking can be modeled using arc-shaped feet at the base of the inverted pendulum \cite{hansen2004roll}. Right: rolling contact CoP defined based on CoM position, resulting in a modified LIP time constant {\cite{hansen2004roll}}.}
    \label{fig:rolling_contact_design}
\end{figure}

The introduction of rolling foot contact, where the CoP goes from the heel to the toe, has been shown to increase the efficiency of walking gaits in biological systems \cite{adamczyk2006advantages}, and can be passively emulated in artificial systems using musculoskeletal models {\cite{narioka20093d}}. While active control of CoP motion in other works has been prescribed by some nominal plan, e.g. \cite{englsberger2014trajectory}, biological data suggests that this heel-to-toe motion is instead a function of CoM position and is invariant to walking speed or time \cite{hansen2004roll}, following the motion of a rolling wheel \cite{mcgeer1990passive} or arced foot \cite{hansen2004roll}, as shown on the left in Fig. \ref{fig:rolling_contact_design}. In this work, we apply this more bio-inspired approach, defining the nominal CoP position only as a function of the CoM position, as in the right of Fig. \ref{fig:rolling_contact_design}.

\begin{figure}[t!]
\centering
    \includegraphics[width=1.0\columnwidth]{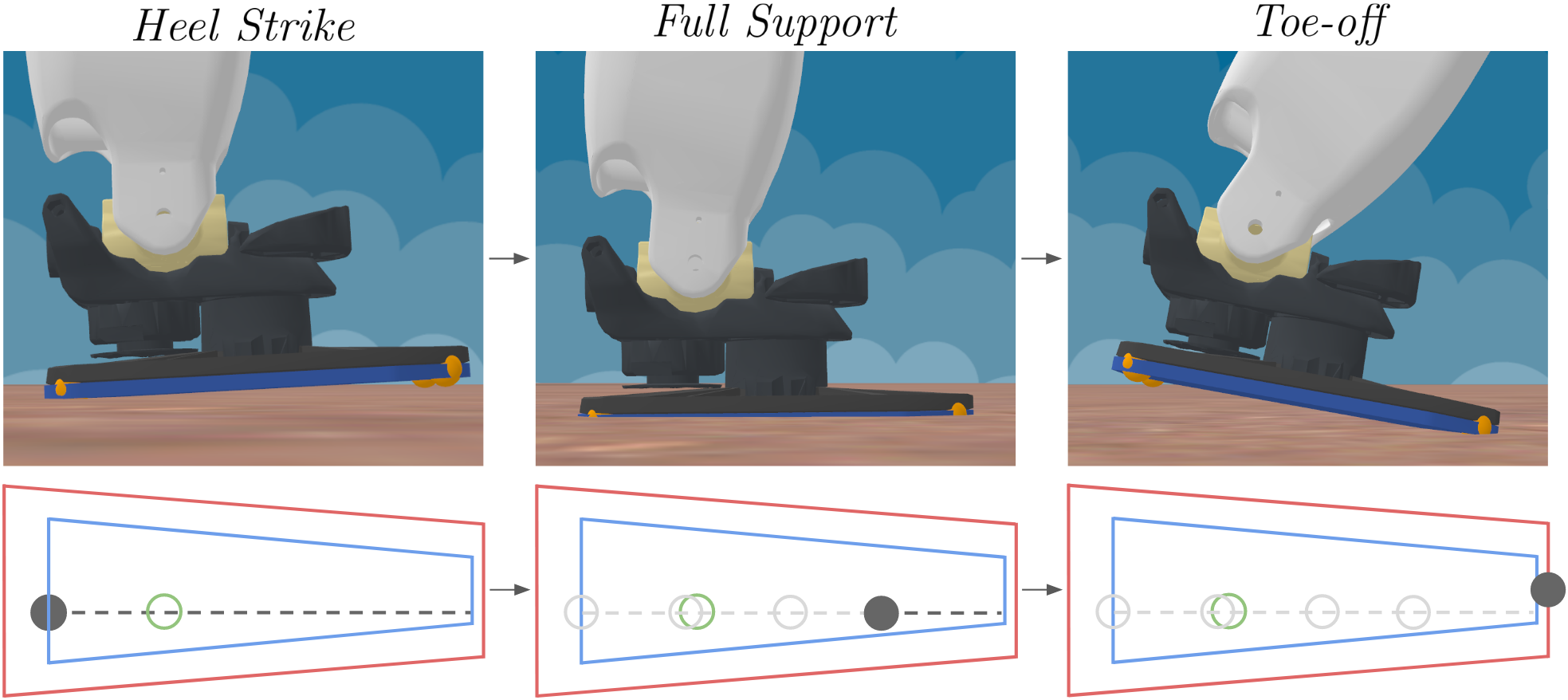}
    \vspace{-4mm}
    \caption{Quickster's full implementation of rolling contact. The non-rolling contact CoP is shown in green. The rolling contact CoP (black) starts at the back of the foot upon heel strike, remains there until toe strike, moves forward toward the toe in full support, and finally reaches and remains at the front of the foot until toe-off.}
    \vspace{-6mm}
    \label{fig:rolling_contact_CoP_travel}
\end{figure}

In the arc-foot rolling contact model, an arc of radius approximately 30\% of the leg length (parameterized by $\alpha$) is added to the foot bottom, which forces the CoP to shift in the direction of the CoM position. When applied to the LIP model, if the CoM is defined relative to the pendulum base, this makes the CoP position relative to the pendulum base 
\begin{equation}
    \mathbf{r}_{cop} = \alpha \mathbf{x},
    \label{eqn:rolling_contact}
\end{equation}
leading to the CoP motion shown in Fig. \ref{fig:rolling_contact_CoP_travel}, where the CoP goes from the heel to the toe as the CoM moves through the gait.
The LIP dynamics from Eq. \ref{eqn:lip_dynamics} then become
\begin{equation}
    \ddot{\mathbf{x}} = \omega^2 \left( \mathbf{x} - \alpha \mathbf{x} \right) \equiv \omega^2 \left( 1 - \alpha \right) \mathbf{x}.
    \label{eqn:rolling_contact_dynamics}
\end{equation}
This is equivalent to the system behaving with a time constant of $\omega \sqrt{1 - \alpha}$, which, in the context of the LIP, means it behaves as a pendulum of height $\frac{\Delta z}{1 - \alpha}$. 
Thus, as this pendulum fraction $\alpha$ for rolling contact gets larger (signifying a larger foot), ALIP behaves as a longer pendulum. The pendulum fraction $\alpha$ used in the Quickster controller at the time of this writing is 0.6.

In biological data, one of the primary energetic benefits of the arc-foot model comes from reduced losses due to impact at the step-to-step transition \cite{adamczyk2006advantages}. 
As the LIP and ALIP model of walking prescribes no impact losses, this benefit is lost. 
However, the use of the rolling contact model applied in Eq. \ref{eqn:rolling_contact_dynamics} reduces the horizontal ground reaction forces by $\left( 1 - \alpha \right)$. 
When the robot is walking forward, it has naturally cyclical forward velocity, where the reaction forces slow and then accelerate the robot from stride to stride. 
Rolling contact should, theoretically, reduce the ``braking" forces applied during each step (reducing negative work), and require decreased ``pushing" forces (positive work) to re-accelerate the mass.

The rolling contact CoP position is then used as an objective for the inverse dynamics-based whole-body controller. 
This, combined with spatial accelerations for the swing foot, chest orientation, and pelvis orientation, desired knee accelerations from the height control policy, and desired arm positions, can be used to find desired torques to be executed by the robot's joints \cite{Koolen_2016}.

%% file: results.tex
To evaluate the efficacy of the Quickster controller in achieving faster, more efficient walking, we conducted several experiments in simulation and on hardware. The baseline controller against which Quickster was compared is the primary walking controller used at IHMC. Unlike Quickster, it utilizes an ankle-based strategy to achieve specific foot placements as determined by a desired CoM trajectory {\cite{griffin2023reachability}}.

\begin{figure*}[!t]
\centering
\includegraphics[width=0.99\textwidth]{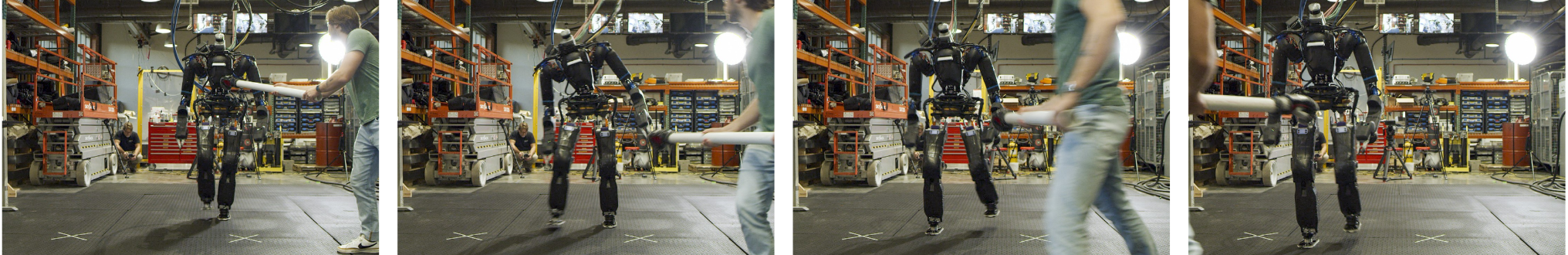}
    \caption{Push recovery while walking forward. The robot adjusts laterally and backwards, and then continues walking. }
    \vspace{-6mm}
    \label{fig:push_recovery}
\end{figure*}

\subsection{Satisfaction of Basic Performance Requirements}

One of the primary goals of the proposed control strategy is to achieve fast, reliable forward walking that exceeds the speeds capable of our baseline walking controller. Extensive testing and development of this controller resulted in reliable forward walking up to 1.4 m/s in simulation and 0.75 m/s on hardware, representing a roughly 40-50\% increase in top speed relative to the baseline. We believe that this improvement can be primarily attributed to the use of a step adjustment-based walking strategy in place of an ankle-based strategy as it provides more control authority over the CoM at higher speeds, while also enabling the exploitation of passive system dynamics to achieve a walking speed rather than strictly relying on step size or duration. While forward walking performance was prioritized during the development of this algorithm, in order to make it as usable as possible, it must also be capable of backwards walking, side-stepping, and turning in place. While no formal speed or efficiency analysis was conducted for these maneuvers, the controller proved capable of performing them reliably throughout numerous tests, both in simulation and on hardware. To assess Quickster's robustness, simulations with blind, unpredictable drops in terrain height were conducted at walking speeds of 0.7 m/s, which the controller was capable of traversing. We also conducted disturbance recovery tests on hardware in which the robot was pushed from all directions while walking in place, and while walking forward as shown in Fig. \ref{fig:push_recovery}. The algorithm proved robust to these perturbations despite its lack of dedicated push recovery functionality, deploying step adjustment to return to a stable limit cycle.

\begin{figure}[!b]
    \centering
    \vspace{-3mm}
    \begin{subfigure}[b]{0.93\columnwidth}
        \centering
        \includegraphics[width=\columnwidth]{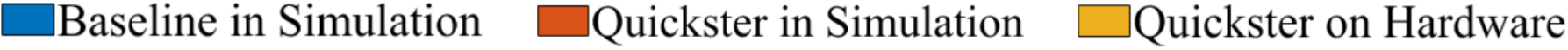}
        \label{fig:rms_power_legend}
        \vspace{-3mm}
    \end{subfigure}

     \begin{subfigure}[b]{0.49\columnwidth}
         \centering
         \includegraphics[width=\columnwidth]{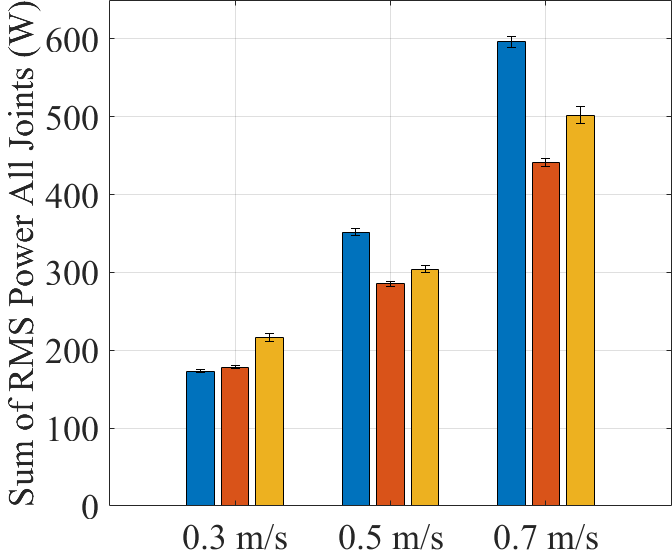}
         \caption{RMS power of all joints}
         \label{fig:rms_power_all_joint}
     \end{subfigure}
     \hfill
     \begin{subfigure}[b]{0.49\columnwidth}
         \centering
         \includegraphics[width=\columnwidth]{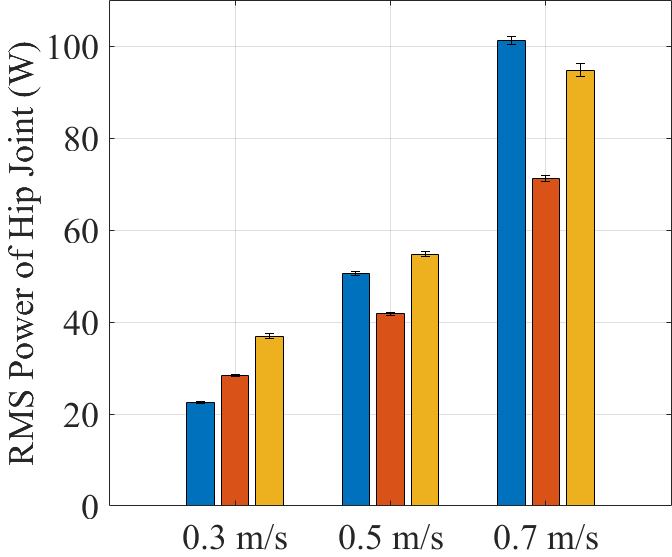}
         \caption{RMS power of hip joint}
         \label{fig:rms_power_hip}
     \end{subfigure}

     \begin{subfigure}[b]{0.49\columnwidth}
         \centering
         \includegraphics[width=\columnwidth]{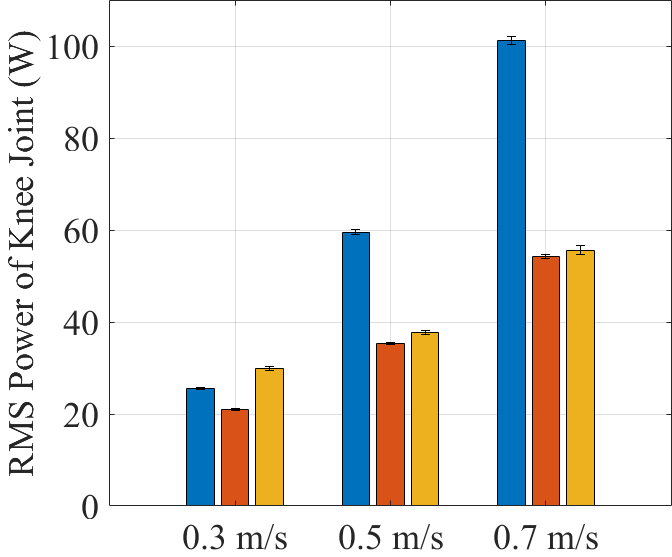}
         \caption{RMS power of knee joint}
         \label{fig:rms_power_knee}
     \end{subfigure}
     \hfill
     \begin{subfigure}[b]{0.49\columnwidth}
         \centering
         \includegraphics[width=\columnwidth]{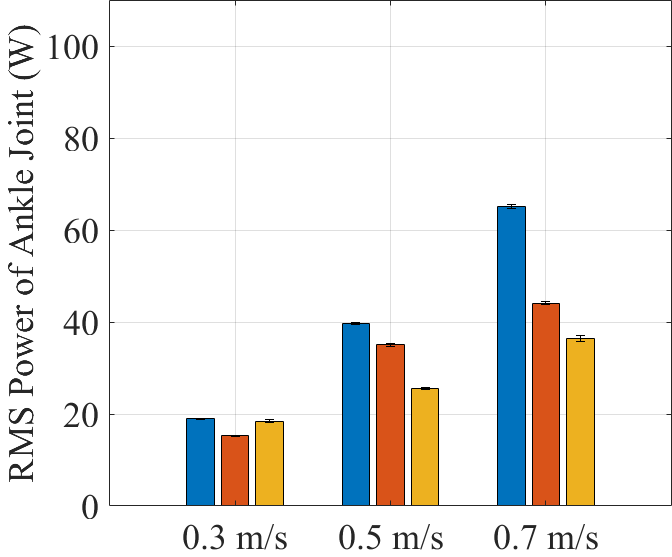}
         \caption{RMS power of ankle joint}
         \label{fig:rms_power_ankle}
     \end{subfigure}
    
     \caption{Comparison of RMS mechanical power expenditure at three steady-state walking speeds.}
     \label{fig:rms_power_comparison}
\end{figure}

\subsection{Gait Efficiency Comparison Against Baseline}
In addition to higher walking speeds, it was our hope that Quickster could also produce an efficient walking gait by considering the angular momentum as part of the step placement dynamics, walking with straighter legs, and utilizing rolling contact in the stance foot. To evaluate this efficiency, we compared the root-mean-squared (RMS) mechanical power of Nadia's leg pitch joints (hip, knee, and ankle) as well as the sum total RMS mechanical power of all joints during steady-state walking in simulation at three different speeds (0.3 m/s, 0.5 m/s, 0.7 m/s) when using Quickster against those measured when using our baseline controller. These speeds were chosen such that the comparison included the slowest walking speed nominally used by the baseline controller, as well as the fastest walking speed that the baseline controller can reliably achieve. The primary comparison was conducted in simulation in order to create the most controlled environment with the least amount of confounding factors or hardware-related complications. However, preliminary hardware experiments were conducted to serve as a reference for the consistency between Quickster's hardware and simulation-based data. 

The results of this experiment can be seen in Fig. \ref{fig:rms_power_comparison}. For nearly all walking speeds and all joints evaluated, the simulation case of the Quickster controller (shown in red) resulted in less RMS mechanical power expenditure when compared against the simulation case of the baseline controller (shown in blue). These reductions ranged from 11.6\% (ankle joint at 0.5 m/s steady-state) to 46.5\% (knee joint at 0.7 m/s steady-state), and tended to increase as walking speed increased. This finding supports our hypothesis that Quickster is not only capable of walking at speeds faster than the baseline, but it also becomes increasingly more efficient in comparison as walking speed increases. There are, however, a few exceptions in which the baseline controller proved more efficient. Firstly, at 0.3 m/s the sum total RMS mechanical power of all joints was slightly higher (2.96\%) when using Quickster compared with that measured when using the baseline. Secondly, Quickster had higher (25.9\%) hip power expenditure than the baseline controller, also at a walking speed of 0.3 m/s. It is very likely that the latter finding (higher hip power), is a significant contributor to the former finding (higher total joint power). These increases in RMS mechanical power could be attributed to the speed and quantity of steps required by the Quickster algorithm. While the baseline controller can take fewer and slower steps at slower walking speed thanks to its use of a CoP feedback-based ankle control strategy, Quickster must step with a certain frequency since step placement is the way in which it maintains control authority over the system dynamics. At faster walking speeds, however, both controllers are required to take quicker steps, and Quickster's better consideration of angular momentum along with its use of straighter legs result in a reliably more efficient gate. This experiment also revealed that Quickster's hardware results (shown in yellow) were relatively consistent with what was measured in simulation, implying that the simulation-based comparison between it and the baseline controller could conceivably be translated to hardware with similar outcomes.

\begin{figure}[!b]
     \centering
     \vspace{-5mm}
     \begin{subfigure}[b]{0.49\columnwidth}
         \centering
         \includegraphics[width=\columnwidth]{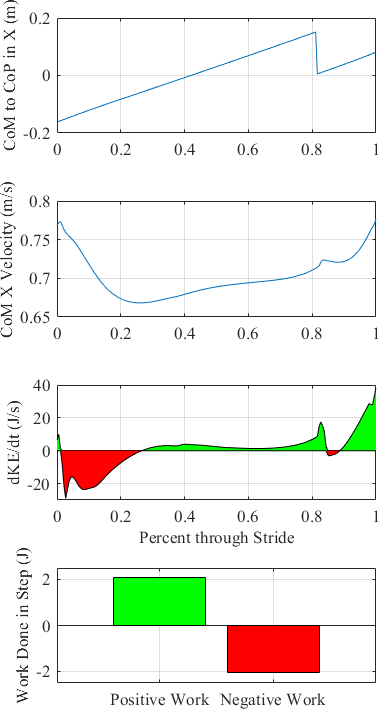}
         \caption{Without Rolling Contact}
         \label{fig:rolling_contact_comparison_without_rolling_contact}
     \end{subfigure}
     \hfill
     \begin{subfigure}[b]{0.49\columnwidth}
         \centering
         \includegraphics[width=\columnwidth]{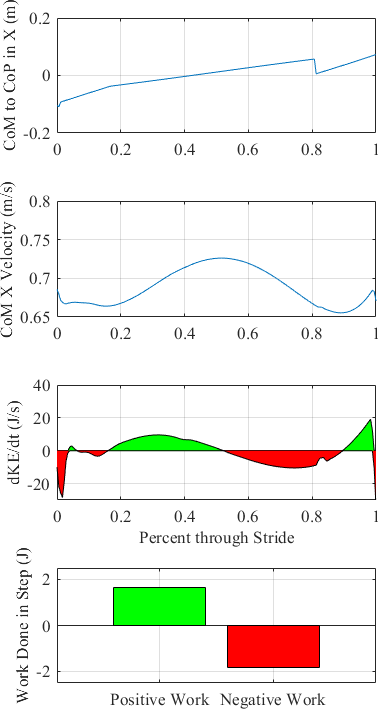}
         \caption{With Rolling Contact}
         \label{fig:rolling_contact_comparison_with_rolling_contact}
     \end{subfigure}
     \vspace{-4mm}
     \caption{Comparison of a single, median step during steady-state walking in simulation at 0.7 m/s with rolling contact turned on (right) and with it turned off (left).}
     \label{fig:rolling_contact_comparison}
\end{figure}

\subsection{Evaluation of Rolling Contact Benefits}

Another potential contribution to the efficiency of Quickster's gait is its use of stance foot rolling contact. To evaluate the effectiveness of this feature, steady-state Quickster walking at 0.7 m/s was conducted in simulation with rolling contact enabled and disabled. A nominal, median steady-state step starting with the touchdown of the current stance foot, and ending with the touchdown of the next stance foot, was chosen to perform this comparison, with results shown in Fig. \ref{fig:rolling_contact_comparison}. The first row of plots shows the CoM position relative to the CoP in the forward ($x$) direction as a function of percent of stride, highlighting the forward motion of the CoM, as well as the sharp decrease in CoM to CoP distance roughly 82\% through the step resulting from toe-off. These profiles, along with active contributions from the controller, dictate the CoM forward velocity which is shown in the second row of plots (relative to world). It can be seen that for both cases, there is a deceleration of the CoM in the beginning of stance corresponding with ground contact, followed by an acceleration as it passes over the CoP towards the end of stance. The aforementioned accelerations and decelerations of the CoM can be correlated to positive and negative work done on the robot at different points throughout the stride. This information can be seen in the third plots which show the time derivative of the CoM's forward kinetic energy as a function of percent of stride. Everything above zero in these plots represents positive work done by the robot (shown in green), and everything below zero represents negative work done on the robot (shown in red). The total amount of positive and negative work done over the course of the step is shown in the bottom bar graphs of Fig. \ref{fig:rolling_contact_comparison}.

This comparison revealed that disabling rolling contact results in a larger CoM deceleration occurring over a longer period of time at the beginning of stride as a result of touchdown. This finding can be attributed to two things. Firstly, in the rolling contact case the location of the CoP at touchdown is closer to the CoM. Secondly, as the CoP rolls forward towards the toe of the stance foot when using rolling contact, the distance between the CoM and CoP decreases in comparison to that same distance when rolling contact is disabled. In both cases, according to LIP dynamics this results in smaller negative and positive CoM accelerations, respectively. However, based on our controller definition, and as can be seen in Fig. \ref{fig:rolling_contact_comparison}, the controller does not behave with the force profile of the LIP. In ALIP dynamics, as is used in our controller, this phenomenon results in a smaller moment arm about the contact point for the gravitational force to act upon, causing less change in the angular momentum. The result of these phenomena is less negative work being done on the robot after touchdown, which translates to less positive work required by the controller to maintain steady-state walking. For this particular median step at this walking speed, using rolling contact provided an 11\% reduction in negative work, and a 21\% reduction in positive work over the course of the stride. These reductions are not equal because both conditions considered are not perfect period one gait cycles due to model and controller non-idealities.

%% file: conclusion.tex
For humanoid robots to be effective, they must be able to quickly and efficiently navigate through their environment. In this work, we sought to develop a controller that enables this skill set. To do so, we designed Quickster; a walking controller that achieves a desired walking speed using ALIP-based step placement, with stability convergence determined using pole-placement. To improve efficiency, we enabled straight-leg walking via direct support leg length control, and implemented a strategy for heel-to-toe rolling contact motion of the stance foot based on biological systems. We then demonstrate that, using this approach, our humanoid robot Nadia is able to stably walk at speeds up to 0.75 m/s on hardware, and recover from external disturbances. Additionally, when compared to our baseline walking controller, the proposed approach shows reductions in RMS joint power expenditure that increase proportionally with walking speed.
Lastly, we show that the use of rolling contact provides a substantial reduction in the work required for decelerating/accelerating the CoM each step.


Several possible extensions exist for future investigation. An improved toe-off behavior and a dedicated arm swinging strategy {\cite{chen2023integrable}} may provide further benefits in the areas of efficiency and angular momentum regulation \cite{kuo2005energetic, meinders1998role, collins2009dynamic}. We are also working on a simple, short-horizon online MPC policy for determination of touchdown positions {\cite{gibson2022terrain}}, a feature that should improve desired walking speed tracking and diminish the negative consequences of model inaccuracies. Lastly, we are working on enabling seamless, autonomous transitions between this controller and our existing walking controller based on environmental models {\cite{Mishra_2021}}, such that Nadia can traverse its environment in the most effective way.